\documentclass{styles/svproc}
\usepackage{graphicx}
\usepackage{multicol}
\usepackage{footmisc}
\usepackage[]{subfigure}
\usepackage{amsmath}
\usepackage{bm}

\usepackage{url}

\begin{document}
\mainmatter              
\title{Mobility Strategy of Multi-Limbed Climbing Robots for Asteroid Exploration}
\titlerunning{Mobility Strategy}  
%
\author{Warley F. R. Ribeiro\inst{1} \and Kentaro Uno\inst{1} \and Masazumi Imai\inst{1} \and Koki Murase\inst{1} \and Bar{\i}\c{s} Can Yal\c{c}{\i}n\inst{2} \and Matteo El Hariry\inst{2} \and Miguel A. Olivares-Mendez\inst{2} \and Kazuya Yoshida\inst{1}}
\authorrunning{Warley F. R. Ribeiro et al.} 
%
\tocauthor{Warley F. R. Ribeiro, Kentaro Uno, Masazumi Imai, Koki Murase, Baris Can Yalcin, Matteo El Hariry, Miguel A. Olivares-Mendez, and Kazuya Yoshida}
\institute{Department of Aerospace Engineering, Graduate School of Engineering, \\ Tohoku University, Sendai, Japan, \\
\email{warley@dc.tohoku.ac.jp},\\
\and
Space Robotics Research Group (SpaceR), SnT-University of Luxembourg, Campus Kirchberg 29, Avenue John F. Kennedy L-1855 Luxembourg}

\maketitle              

\begin{abstract}
Mobility on asteroids by multi-limbed climbing robots is expected to achieve our exploration goals in such challenging environments. We propose a mobility strategy to improve the locomotion safety of climbing robots in such harsh environments that picture extremely low gravity and highly uneven terrain. Our method plans the gait by decoupling the base and limbs' movements and adjusting the main body pose to avoid ground collisions. The proposed approach includes a motion planning that reduces the reactions generated by the robot's movement by optimizing the swinging trajectory and distributing the momentum. Lower motion reactions decrease the pulling forces on the grippers, avoiding the slippage and flotation of the robot. Dynamic simulations and experiments demonstrate that the proposed method could improve the robot's mobility on the surface of asteroids.

\end{abstract}

\section{Introduction}

Asteroids have been one of the many celestial bodies receiving attention in space exploration activities, mainly due to their potential to unlock scientific, commercial, and safety benefits. They are relatively small rocky bodies orbiting our Sun, composed of water, metals, and other rare materials, making them appealing for scientific missions and space mining.
Asteroids could also be a threat to our planet due to impact hazards. Having the ability to prevent such risk is a technology we need to keep our home planet safe~\cite{Badescu2013}.

A deep understanding of the nature and detailed composition of asteroids is needed to achieve our objectives, and it is necessary to conduct in-situ exploration and observation missions. Robotic exploration is effective and efficient for such hazardous environments for humans, as demonstrated by previous and current space exploration missions to the Moon, Mars, and other celestial bodies~\cite{yoshida2008}.

Space agencies conducted many missions to orbit or to fly by asteroids to study their characteristics. The Japanese missions Hayabusa and Hayabusa2 were the first to collect and return surface samples to Earth~\cite{yano2006touchdown,watanabe2017hayabusa2}. NASA also performed the first planetary defense test with the DART mission, changing the trajectory of a small asteroid through kinetic impact~\cite{reichert2023dart}.

Mobile robots would be necessary for a more extensive in-situ exploration of asteroids, but these small celestial bodies present some unique challenges. The low gravity, usually more than a thousand times smaller than Earth's, means that traditional mobility methods are unsuitable on the surface of asteroids. The lack of atmosphere of these small bodies makes air-based locomotion systems unfeasible. Additionally, extremely rough and unknown terrain shapes with fragile and porous structures present additional challenges for surface mobility~\cite{michikami2021boulder}.

\subsection{Related Works}

Different studies have investigated the use of alternative mobility systems to explore asteroids. Hopping robots is one creative option, as it uses low gravity in its favor to generate a jumping motion with small mechanisms. The MINERVA-II-1 robots aboard the Hayabusa2 mission performed the first surface locomotion exploration on an asteroid, demonstrating the feasibility of the hopping method in actual exploration missions~\cite{yoshimitsu2022asteroid}. However, once the robot jumps, it can rebound in an unpredictable direction due to the uneven nature of the terrain. 

\begin{figure}[t]
\centering
\includegraphics[width=0.9\textwidth]{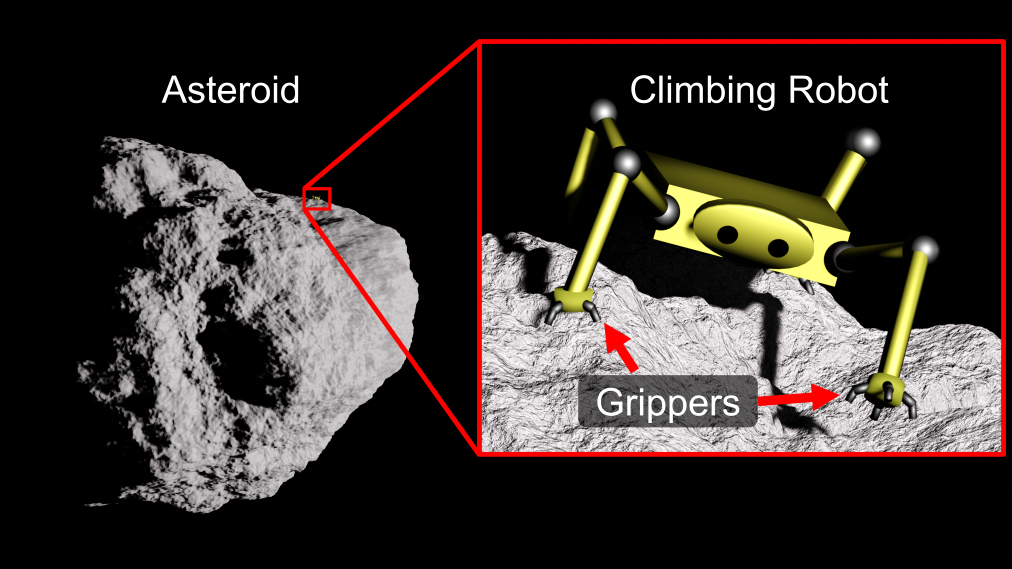}
\caption{Concept image of a climbing robot exploring the surface of an asteroid.}
\label{fig:legged}
\end{figure}

Multi-legged climbing robots were also proposed as a locomotion solution for asteroid exploration missions, as pictured in the conceptual image in Fig.~\ref{fig:legged}~\cite{yoshida2002novel}. Although walking robots are more complex and heavier due to the multiple actuators, the possibility of achieving precise mobility is sufficient to attempt this solution for asteroid exploration. Moreover, the legs (or more precisely, the limbs) of the robot can also act as manipulators to handle objects, such as surface samples or tools for analyses.
Adding spine-type grippers to the robot can improve its stability, enhancing its motion capability to avoid undesired flotation. NASA JPL proposed a gripper with multiple microspines to be integrated into the LEMUR series robots for asteroid exploration~\cite{parness2013gravity}.

The locomotion planning of multi-limbed robots for asteroid exploration needs to consider the effects of the low gravity to perform mobility without flotation failures caused by the grippers' detachment. The quasi-static analysis is a general solution for slow movements~\cite{parness2017lemur}, but even low-speed motions could cause the grippers to detach. Another proposed method considers equilibrium conditions from the Zero Moment Point (ZMP) and estimation of contact forces to plan the robot's motion~\cite{chacin2009motion}. However, estimating contact forces in unknown environments could be challenging, and ZMP analysis is not trivial for uneven terrains. As another alternative for mobility in microgravity, reactionless control was proposed for bipedal climbing robots~\cite{yuguchi2016analysis}. However, this method did not account for multiple contact points and showed problems with kinematic singularities.

\subsection{Objective and Contributions}

Despite the progress in robotic mobility for asteroid exploration, a significant opportunity for further improvement and development remains. This study aims to propose an alternative for limbed robot locomotion on asteroids that addresses the limitations of current methods while being capable of moving precisely in rough terrains.

Traversing over a rough surface requires planning the footholds and base posture that generates feasible configurations for the robot. Mobility in microgravity also requires planning to reduce the reactions induced by the robot's motions to prevent flotation.

We propose a locomotion strategy for asteroid exploration that combines gait and motion planning for climbing robots to achieve safe mobility on rough surfaces under microgravity. The gait planning selects the base pose that provides a feasible motion based on the limbs' contact points on the uneven surface. The motion planning includes reaction-aware techniques to enhance locomotion safety in microgravity to avoid the robot's flotation, keeping precise and feasible mobility.

The main contributions of this paper are the following:

\begin{enumerate}
    \item Locomotion strategy that reaches a precise location on an asteroid's surface, lowering risks of mobility failure. This strategy includes gait planning that selects a feasible robot's posture, and motion planning that reduces the induced motion reactions.
    \item Validation of the proposed mobility strategy by dynamic simulations of a multi-limbed robot moving on a rough surface and experiments with a microgravity emulating facility.
\end{enumerate}

First, we present the mobility strategy, presenting the techniques to select footholds and base posture to achieve feasible motions. We also introduce motion planning that reduces the flotation risk by using reaction-aware methods to minimize the motion reactions. Then, we present simulation and experimental studies to validate the proposed methods.

\section{Mobility Strategy in Microgravity}

Unlike Earth-based walking robots, locomotion on an asteroid's surface requires additional or alternative approaches to achieve safe mobility. The lack of gravity makes it necessary for a robot to control its movements carefully. Otherwise, it could generate pulling forces, causing the slippage of the grippers in contact with the ground. The significantly rough surface requires the robot to constantly adapt its posture to achieve feasible configurations as the limbs contact the surface at different heights.

\subsection{Gait Planning}

Planning the sequence of steps and movements of a limbed robot requires defining which limb is moving at what time, where should its landing position be, and how to conduct the robot's base motion. Usually, climbing robots require a more complex strategy, as the terrain is uneven and the available graspable points are sparsely distributed.

\begin{figure}[t]
\centering
\includegraphics[width=0.99\textwidth]{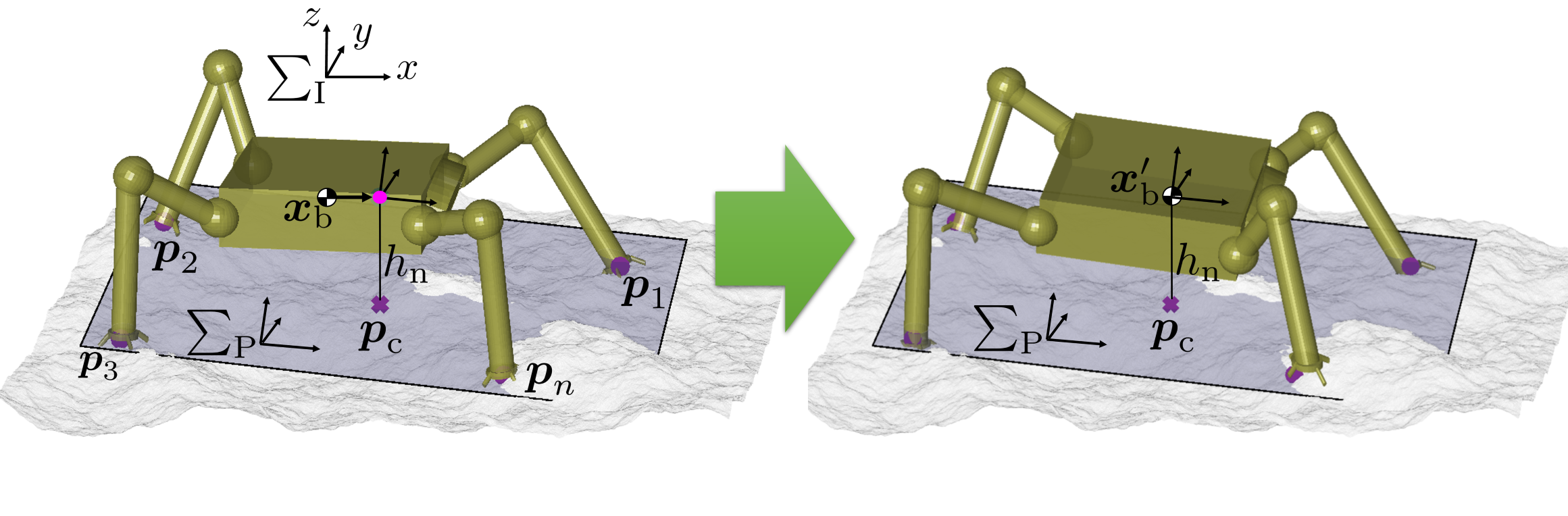}
\caption{Scheme for the desired motion of the robot's main body, based on the linear regression plane of the contact points.}
\label{fig:base_pose}
\end{figure}

In general, the gaits of climbing robots are non-periodic, allowing the robot to adjust to the discrete points' distance and availability around each given limb. The selection of the next grasping point can also follow different criteria, such as the robot's stability, the desired moving direction, and the feasible workspace~\cite{uno2019gait}. For asteroid exploration robots, the gripper's capabilities and terrain shape define if the graspable points are discrete. In this paper, for simplification purposes, we assume that the robot can grasp any location on the surface, assuming the robot has a mechanism capable of attaching to concave and convex shapes~\cite{kato2022pin}. However, any leg sequence should be applicable to the strategy presented here as long as it ensures feasible motions.

As for the base movement, we separate the base and limb swinging motion, allowing the base motion to be used during the swinging phase, as shown later in the motion planning strategy. Therefore, after each step of the robot, a base adjustment is included during the supporting phase. This work considers that the swinging and supporting periods are equal, producing a total period $T=2nT_{\text{sw}}$, where $n$ is the number of limbs, and $T_{\text{sw}}$ is the swinging period.

We define the desired base pose $\bm{x}_{b}'~\in~\bm{R}^6$ for each supporting phase from the supporting polygon defined from the contact points. As shown in Fig.~\ref{fig:base_pose}, the supporting polygon is the linear regression plane of all the contact points $\bm{p}_1$, $\bm{p}_2$, $\ldots$ , $\bm{p}_n$~\cite{chen2022adaptive}. The centroid position $\bm{p}_c~\in~\bm{R}^3$ of the contact points defines the horizontal components $x$ and $y$ for the desired base position. The base height ${x}_{b_z}'$ is the nominal stance height $h_{\text{n}}$ added to the supporting polygon vertical position at the centroidal location $\bm{p}_{c_z}$. The desired attitude of the robot is the same as the orientation of the supporting plane, defined by the regression plane's coordinate frame $\Sigma_p$.

With the robot's base pose defined by the regression plane of contact points, the robot stays closer to its nominal stance, regardless of the shape of the terrain it is traversing. However, collisions between the base and the surface could still happen if the ground presents obstacles on the path the main body moves. Therefore, we include a base collision avoidance strategy, assuming the terrain information is available to the robot's planner. An additional elevation is added to the base height if the planner detects a collision for the desired base pose, i.e., the vertical coordinate of the point on the inferior base plane is smaller than the respective coordinate of the surface.  We define the desired base height to avoid ground collisions in~(\ref{eq:z_coll}), by adding the maximum detected collision depth $d_{\text{coll}}$, an additional height $h_{\text{add}}$ to guarantee ground clearance, and subtracting the size $b_\text{{low}}$ of the lower portion of the base body.

\begin{equation}
    {x}_{b_z}' = {p}_{c_z} + h_{\text{n}} + \max(\left| d_{\text{coll}} \right|) + h_{\text{add}} - b_\text{{low}} 
    \label{eq:z_coll}
\end{equation}

\subsection{Motion Planning}

Once we decide the desired stance of each step, including the desired positions of all limbs and the base's pose, the climbing robot needs to perform the motion of its joints to reach the selected configuration. Assuming a multi-body system with multiple rigid links connected by rotational joints, the planner has to find a feasible movement for the actuators to achieve the desired stance. Moreover, considering the lack of gravity on the surface of asteroids, the planned motion needs to avoid excessive motion reactions that could provoke the slippage of the attached grippers.

\begin{figure}[t]
\centering
\includegraphics[width=0.55\textwidth]{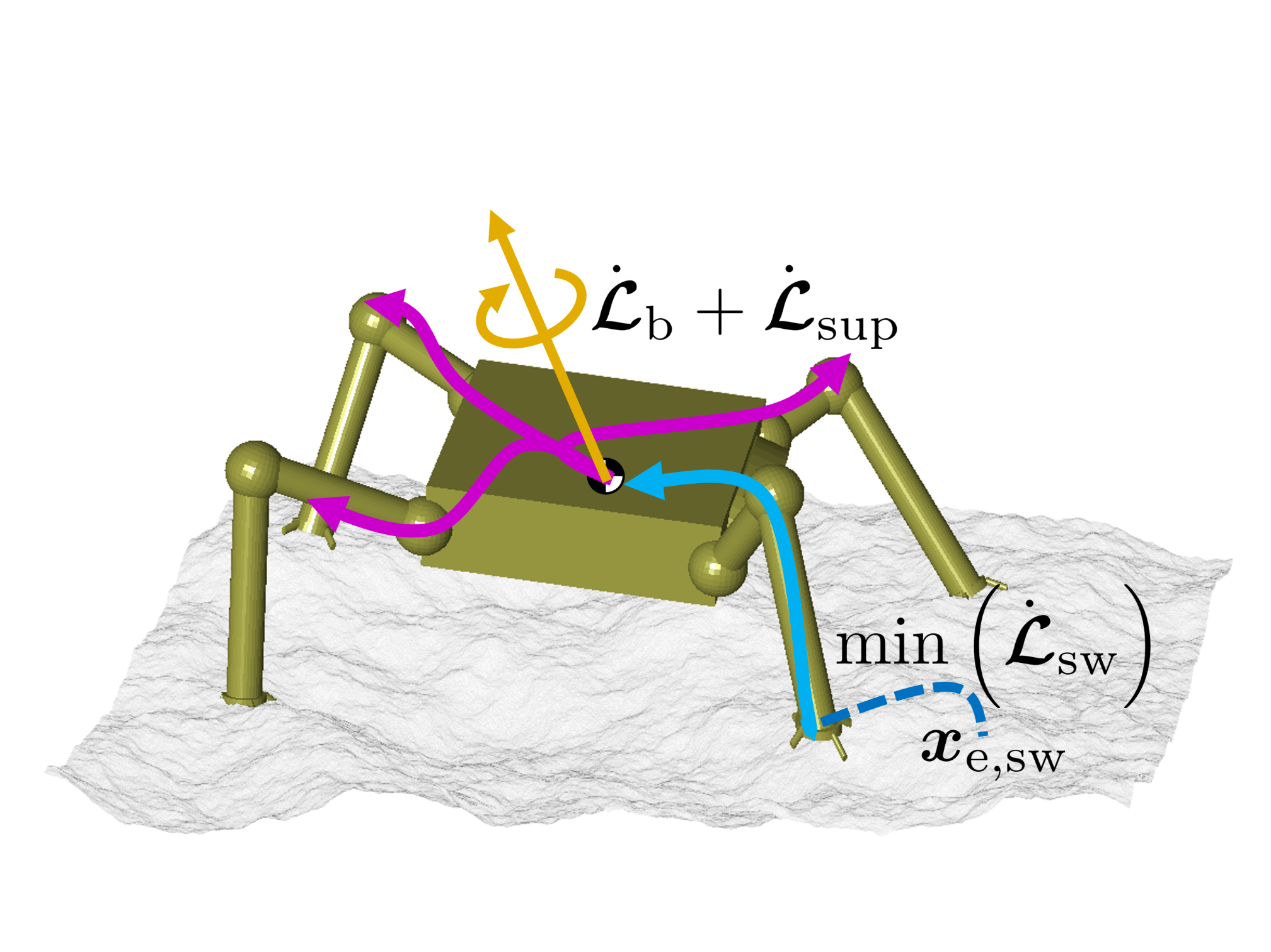}
\caption{Concept image of the Reaction-Aware Motion Planning.}
\label{fig:RAMP}
\end{figure}

We previously proposed Reaction-Aware Motion Planning (RAMP), which reduces the total change of linear and angular momentum of the intended motion, improving mobility safety in microgravity~\cite{ribeiro2023ramp}. Fig.~\ref{fig:RAMP} shows a concept image of RAMP, divided into two strategies to reduce motion reactions. The first is the Low-Reaction Swing Trajectory (LRST), describing the trajectory planning for the swinging limbs that lowers the motion reactions~\cite{ribeiro2022low}. The second is the Momentum Distribution (MD), representing the compensation of reactions generated by the swinging motion with the swinging momentum distribution to other parts of the robot, i.e., the supporting limbs and the robot's main body.

The LRST optimizes the coefficients $\boldsymbol{A}_{Bj}$ of a polynomial Bezier curve that connects the current and desired positions of the swinging limb, minimizing the variation of linear and angular momentum, $\boldsymbol{\dot{\mathcal{L}}}_\text{lin}$ and $\boldsymbol{\dot{\mathcal{L}}}_\text{ang}$. As described in~(\ref{eq:LRST}), LRST also ensures the feasibility of the swinging motion by the inequalities that guarantee the joint angles $\boldsymbol{\phi}_{\textrm{sw}}$ are within the physical limits of the robot. The objective function LRST also provides a term to increase the step height to a predefined value $h_\text{sw}$ to avoid ground collisions of the gripper during the swinging phase. The remaining coefficients of the Bezier curve are computed from the boundary conditions, assuming zero velocity and acceleration at both initial and final times, $t_0$ and $t_f$. The coefficients $C_1$, $C_2$, and $C_3$ define the weight of each term in the objective function.

\begin{equation}
    \begin{aligned}
        \min_{\boldsymbol{A}_{B3},\boldsymbol{A}_{B4}} \quad  & {C}_1 \max \left( | \dot{\boldsymbol{\mathcal{L}}}_\text{lin} \left( t \right) | \right) + {C}_2 \max \left( |\dot{\boldsymbol{\mathcal{L}}}_\text{ang} \left( t \right) | \right) + C_3 \left| h_\text{sw} - \max \left( x_{\text{e,sw}_z} \left( t \right) \right) \right|    \\[1.8ex]
        \textrm{s. t.} \quad 
            & \boldsymbol{x}_{e,\text{sw}}(t) = \sum_{j=0}^{7}  \boldsymbol{A}_{Bj} \begin{pmatrix} 7 \\ j \end{pmatrix} \left( \frac{t_f - t}{t_f - t_0} \right)^{7-j}  \left( \frac{t - t_0}{t_f - t_0} \right)^{j}  \\
            & \boldsymbol{\phi}_{\textrm{sw},\min} \leq \boldsymbol{\phi}_{\textrm{sw}}(t) \leq \boldsymbol{\phi}_{\textrm{sw},\max} 
    \end{aligned}
\label{eq:LRST}
\end{equation}

While LRST defines the swinging limb trajectory, MD computes the motion of the robot's base. By distributing a fraction $\alpha$ of the momentum generated by the swinging motion to all the remaining links of the robot, we can reduce the total momentum change, i.e., lower the forces generated by the robot that could induce the slippage of the grippers. From the definition of the total momenta of a multi-limbed robot in~(\ref{eq:momentum_base+legs}), we can compute the base velocity during the swinging phase in~(\ref{eq:momentum_distribution}). Here, $\boldsymbol{H}_{b}$ and $\boldsymbol{H}_{bm,i}$ represent the inertia of the robot and the $i$-th limb, and $\boldsymbol{J}_{b,i}$ and $\boldsymbol{J}_{m,i}$ are the Jacobian matrices.

\begin{equation}
  \begin{aligned}
    \boldsymbol{\mathcal{L}} = & \ 
    \boldsymbol{\mathcal{L}}_{\text{b}}  + \boldsymbol{\mathcal{L}}_{\text{sup}}   + \boldsymbol{\mathcal{L}}_{\text{sw}}  \\
    = & \ 
    \boldsymbol{H}_{b} \dot{\boldsymbol{x}}_{b}(t) + 
    \sum_{i=1}^{n_{\text{sup}}} \boldsymbol{H}_{bm,i} \dot{\boldsymbol{\phi}}_i(t) + 
    \sum_{i=1}^{n_{\text{sw}}} \boldsymbol{H}_{bm,i} \dot{\boldsymbol{\phi}}_i(t)
   \end{aligned} 
   \label{eq:momentum_base+legs}
\end{equation}

\begin{equation}    
    \dot{\boldsymbol{x}}_{b}(t) = 
    -\alpha \left( \boldsymbol{H}_{b} - \sum_{i=1}^{n_{\text{sup}}} \boldsymbol{H}_{bm,i} \boldsymbol{J}_{m,i}^{+} \boldsymbol{J}_{b,i} \right)^{-1}  \sum_{i=1}^{n_{\text{sw}}} \boldsymbol{H}_{bm,i} \dot{\boldsymbol{\phi}}_i (t)
    \label{eq:momentum_distribution}
\end{equation}

By numerically integrating the base velocity to obtain the base pose $\boldsymbol{x}_{b}(t)$, we can define the robot's configuration at each instant during the swinging phase. During the supporting phase, when only the main body moves, we generate a simple polynomial trajectory for the base pose until the desired pose, defined in the gait planning to avoid collisions using the linear regression plane.

In our previous work, the operator defined the momentum distribution factor alpha according to the robot's inertial and locomotion parameters. In this paper, we propose an improvement by automatically setting the distribution factor at each time step during the swinging phase. If the robot is close to a kinematic singularity, the distribution factor has to be decreased to avoid mobility failures. Therefore, we propose using the manipulability measurement defined in~(\ref{eq:manipulability}) as an element to specify how much momenta can be distributed from the swinging limb to the remaining parts of the robot.

\begin{equation}    
    w_i(t) = \sqrt{\text{det} \left( \boldsymbol{J}_{m,i}(t) \boldsymbol{J}_{m,i}(t)^T \right)}
    \label{eq:manipulability}
\end{equation}

A simple linear function defines the value of the momentum distribution factor at each instant in~(\ref{eq:distr_factor}) from the maximum and minimum values of manipulability and the minimum manipulability at that time $\min \left( w_i(t) \right)$, considering all possible limbs.

\begin{equation}    
    \alpha(t) = \frac{ \min \left( w_i(t) \right) - w_{\min}} { w_{{\max}} - w_{\min}} 
    \label{eq:distr_factor}
\end{equation}

\section{Simulated Case Studies}

We performed dynamic simulations to validate our proposed mobility strategy for climbing robots on asteroids. ClimbLab is a numerical simulator developed for climbing robots using MATLAB\copyright, allowing rapid implementation of new gait and motion patterns~\cite{uno2022climblab}. For the model of the robot, we chose HubRobo, a quadruped climbing robot developed by our group~\cite{uno2021hubrobo}. And for the environment, a fractal surface with a standard deviation of 30~mm for the terrain elevation and gravity level of 10$^{-6}$~G. The surface contact force is modeled by a compliant model, where the stiffness and damping coefficients are 4000~N/m and 1~Ns/m, respectively. Assuming a  maximum holding force of 0.9~N, the robot's gripper detaches if the contact force exceeds this value in a pulling direction.

The robot walks with a periodic gait, moving first the rear limbs, then the front ones. The total cycle period $T$ is 14~s, leaving each swinging and supporting phase with 1.75~s. With a stride of 8~cm and a step height $h_{\text{sw}}$ of 4~cm, the robot can cover the surface with a velocity of 0.57~cm/s.

\begin{figure}[t]
\centering
\subfigure[Motion without proposed strategy]
{\includegraphics[width=0.65\linewidth]{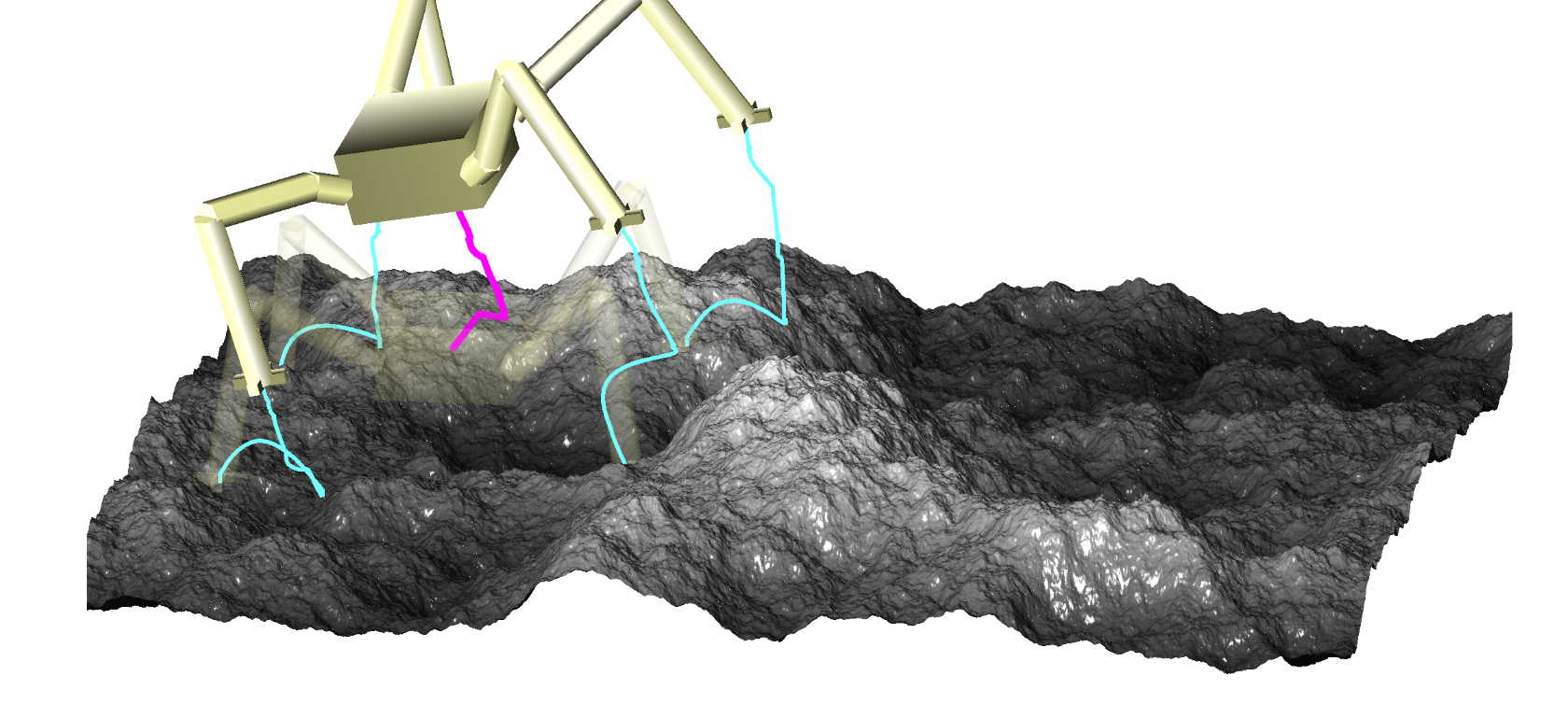}
\label{fig:sim_BL}
}
\hspace{-1mm}
\subfigure[Motion with proposed strategy]
{\includegraphics[width=0.65\linewidth]{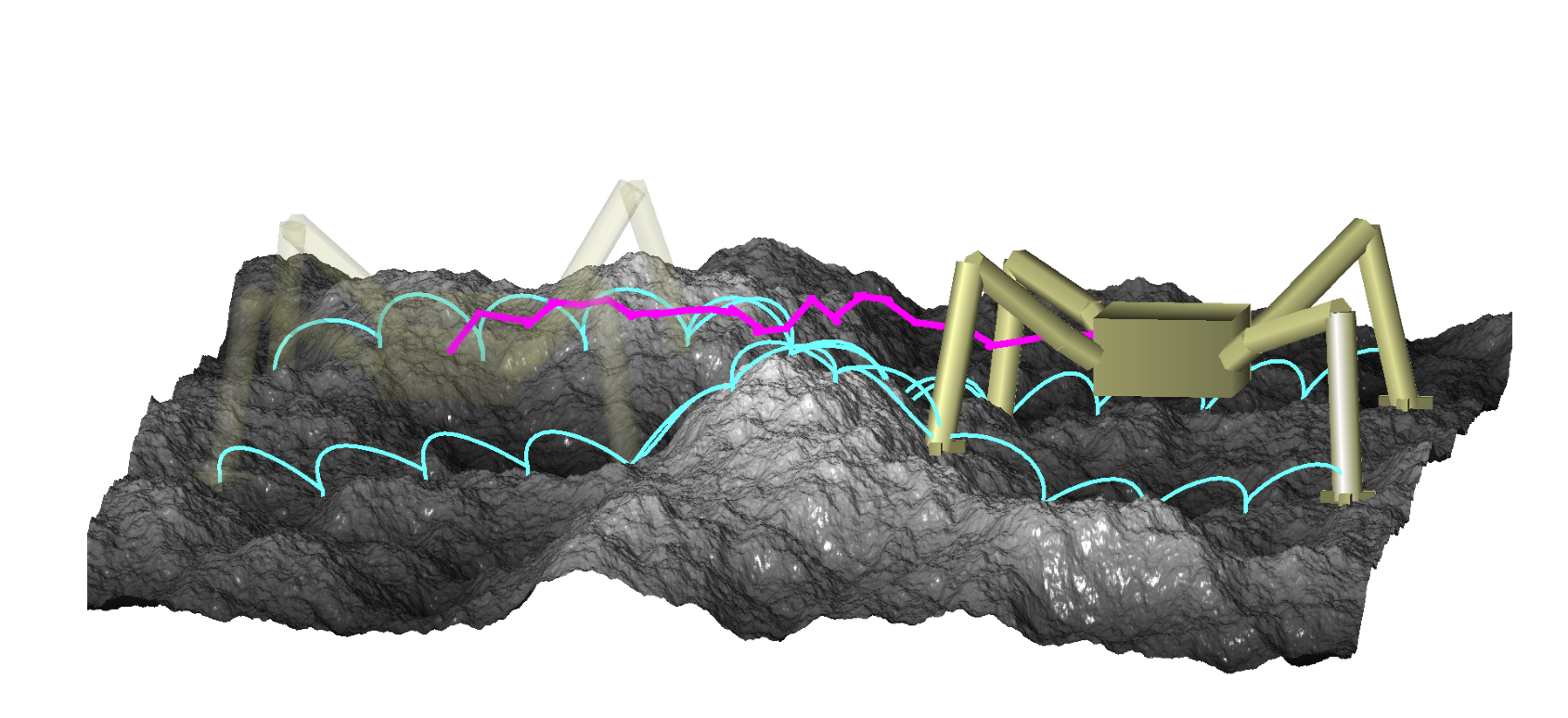}
\label{fig:sim_RAMP}
}\\
\caption{Snapshots of simulations}
\label{fig:sim}
\end{figure}

Fig.~\ref{fig:sim} shows the results of simulations with and without the proposed methods. In the first case, the robot swings its limbs with a regular polynomial trajectory, not moving the base to distribute momentum during the swinging phase and not adjusting the main body to avoid ground collisions. The second simulation implements all strategies proposed in this paper, with an additional height $h_{\text{add}}$ of 2~cm for the ground clearance in case of detected collisions, LRST coefficients $C_1=7$, $C_2=1.75$, and $C_3=30$, and maximum and minimum manipulability of 1.2~mm and 2.5~mm, respectively.

The results show that the robot's grippers detach when moving without the proposed method after walking only a few steps. This result shows how the reactions generated by the dynamic motion of a climbing multi-limbed robot could induce failure on asteroids due to the lack of gravity. The result of locomotion with the proposed method shows two essential improvements for asteroid exploration with climbing robots. The first is the reduction of reactions through RAMP, allowing mobility without grippers' slippage. The second is collision avoidance using the base pose adjustment of the proposed gait planning, preventing impacts that could also provoke detachment and flotation of the robot. Additionally, the robot avoids unfeasible configurations automatically by updating the momentum distribution factor using the manipulability index.

\begin{figure}[!t]
\centering
\subfigure[Max. Contact Force]
{\includegraphics[width=0.48\linewidth]{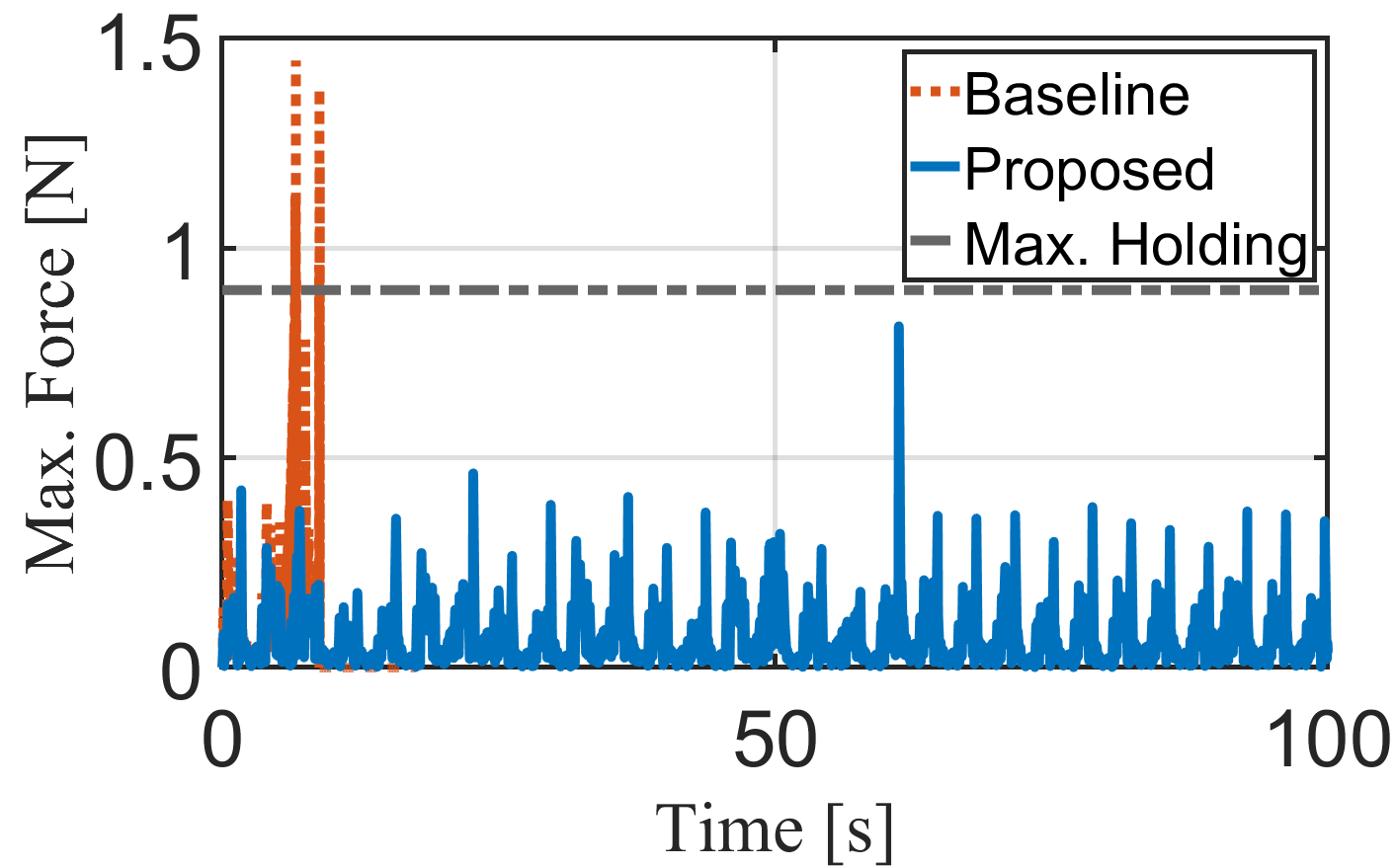}
\label{fig:force}
}
\hspace{-0.5mm}
\subfigure[GIA Margin]
{\includegraphics[width=0.48\linewidth]{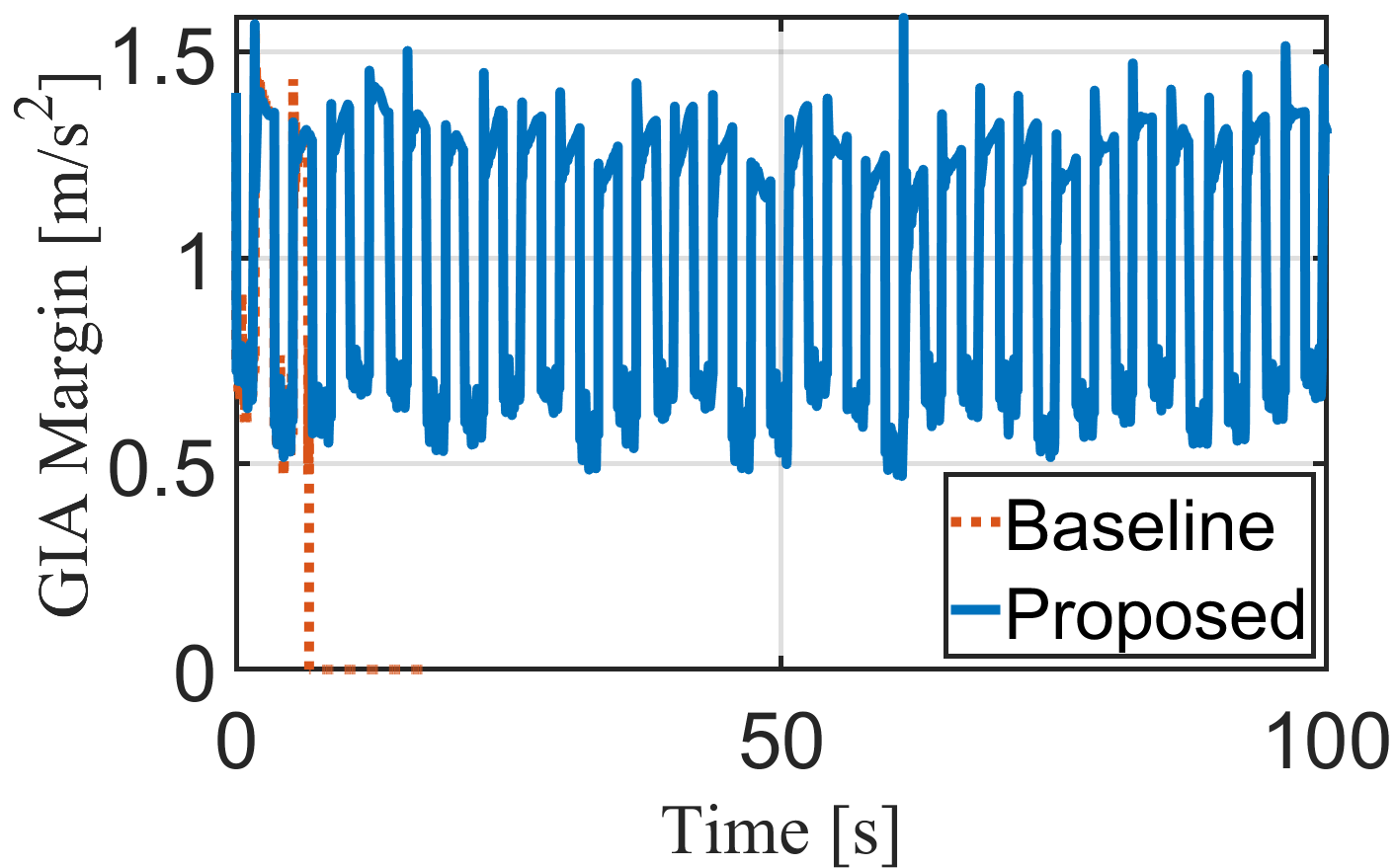}
\label{fig:gia}
}
\caption{Comparison graphs between simulations with and without the proposed strategy for mobility on asteroids.}
\label{fig:sim_graph}
\end{figure}

Fig.~\ref{fig:sim_graph} shows graphs comparing the simulation results of the motions with and without the proposed trajectory. The first graph displays the maximum contact force, considering all possible grippers attached to the ground. The second one shows the Gravito-Inertial Acceleration (GIA) margin, a metric to define the dynamic equilibrium of climbing robots~\cite{ribeiro2020dynamic}. Both results validate the proposed method as a viable solution to reduce motion reactions and increase locomotion safety by decreasing the risk of gripper slippage. In both graphs, we can also observe the moment failure happens for the first simulated case when the maximum contact force overcomes the maximum holding limit of the gripper or the GIA becomes zero. While for the motion with the proposed strategy, detachment does not occur because the GIA margin is always positive, and the ground reaction forces are always under the gripper holding limit.

\section{Experimental Case Study}

Experimental evaluation for microgravity conditions is challenging on Earth, and testing in actual microgravity environments is expensive. We used an air-floating system to emulate microgravity in a two-dimensions planar configuration. A robotic platform mounted with air bearings can levitate to create a thin layer of air between the robot and a flat surface, eliminating friction forces to create an effect similar to microgravity in the horizontal plane~\cite{yalcin2023ultra}. Two robotic arms are mounted on the floating platform to simulate a limbed robot. The robotic arms have pinching grippers capable of grasping fixed structures that emulate the ground surface of an asteroid.

We use the air floating system to test our proposed mobility strategy, comparing it with a regular baseline locomotion method for climbing robots. In both cases, the robot walks with a periodic gait, using 5 s to release the gripper and another 5 s to grasp the new target. Swinging motion and base movement period also take 5 s each, for a period T = 40 s, as the robot has only two limbs. The stride and step-height are 10 cm and 4 cm, respectively. Due to the simple linear arrangement of the grasping targets, the proposed gait planning does not perform any adjustment to avoid collisions. As for motion planning, using RAMP generates a different trajectory and changes the base posture during the swinging phase to minimize motion reactions. The trajectory optimization coefficients are $C_1=40$, $C_2=0$, and $C_3=8$, while the momentum distribution factor is 0.3 throughout the swinging phase as a conservative measure to avoid kinematic singularities.

\begin{figure}[t]
\centering
\subfigure[$t=0$~s]
{\includegraphics[width=0.48\linewidth]{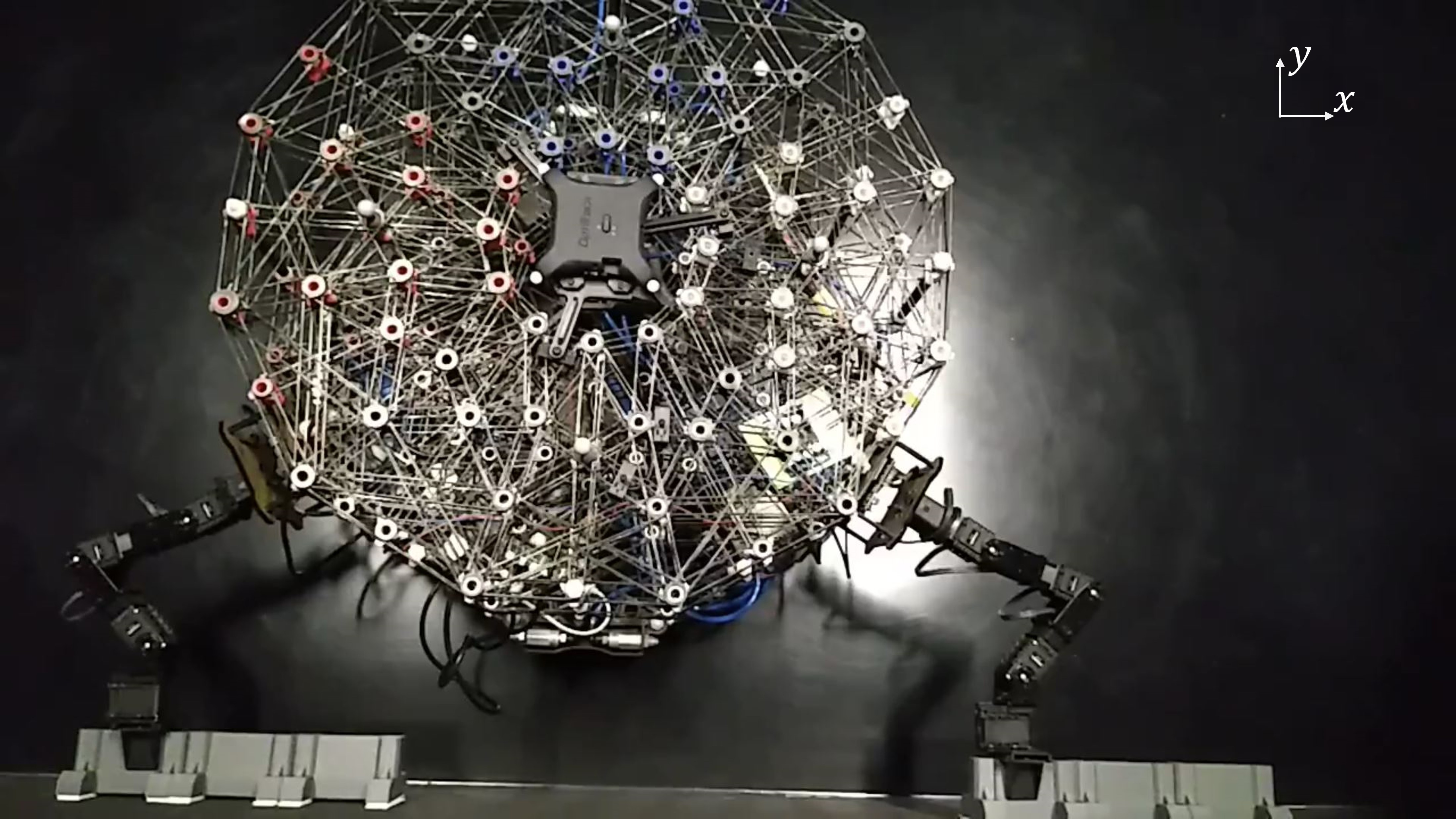}
\label{fig:exp_BL0}
}
\hspace{-1mm}
\subfigure[$t=20$~s]
{\includegraphics[width=0.48\linewidth]{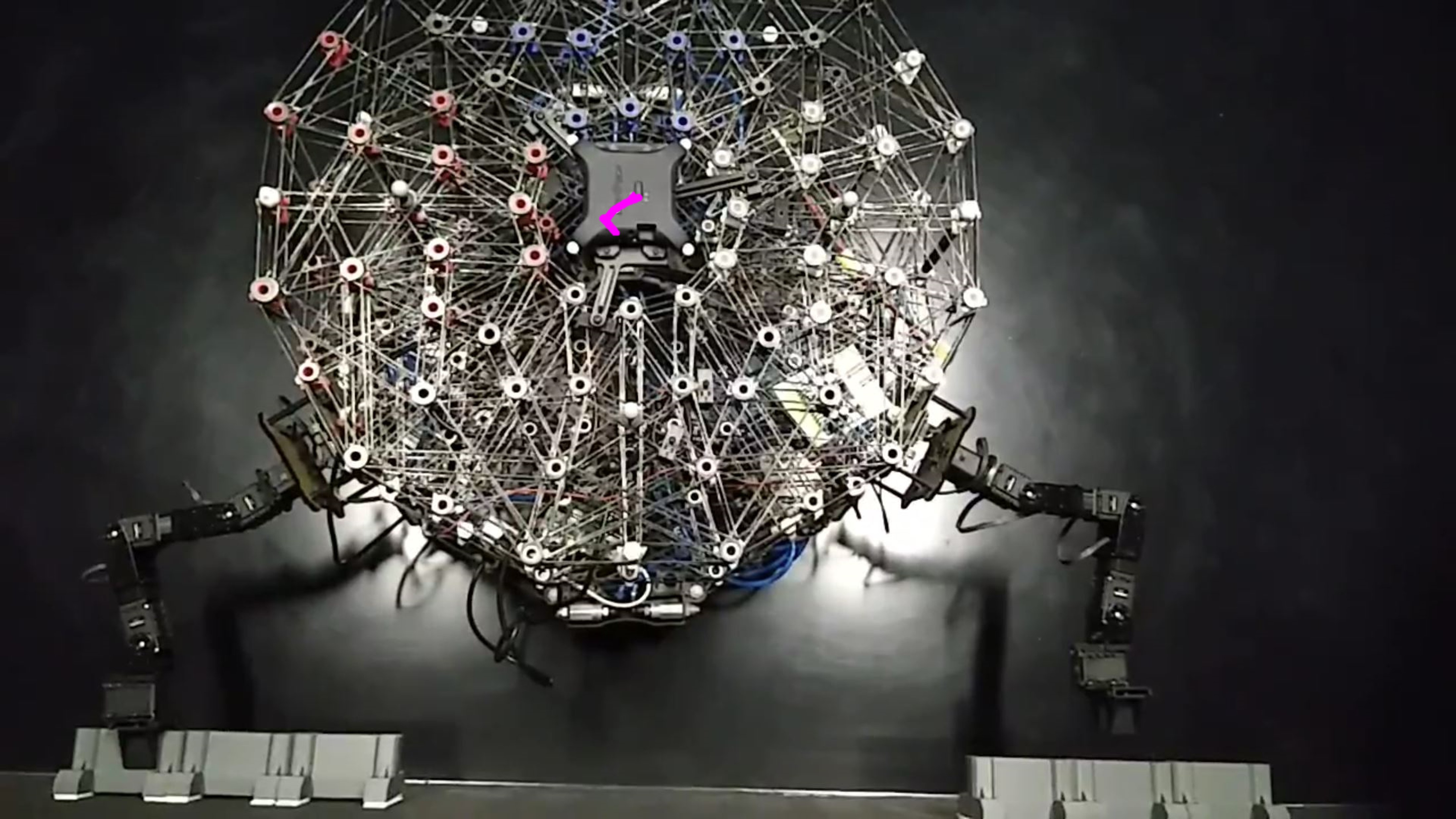}
\label{fig:exp_BL20}
}\\
\vspace{-2.5mm}
\subfigure[$t=28$~s]
{\includegraphics[width=0.48\linewidth]{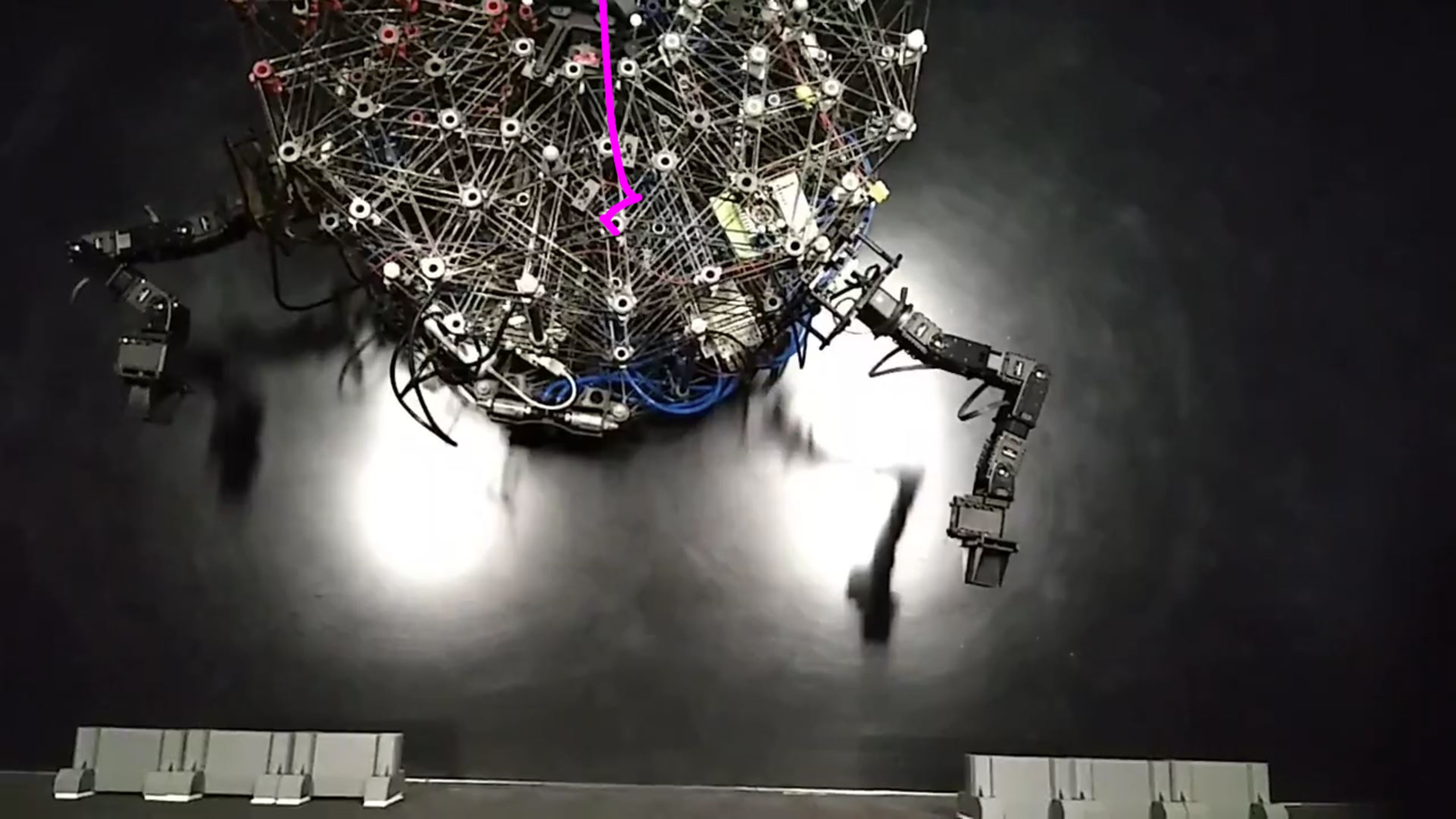}
\label{fig:exp_BL28}
}
\caption{Snapshots of experiments}
\label{fig:exp_BL}
\end{figure}

\begin{figure}[t]
\centering
\subfigure[$t=0$~s]
{\includegraphics[width=0.48\linewidth]{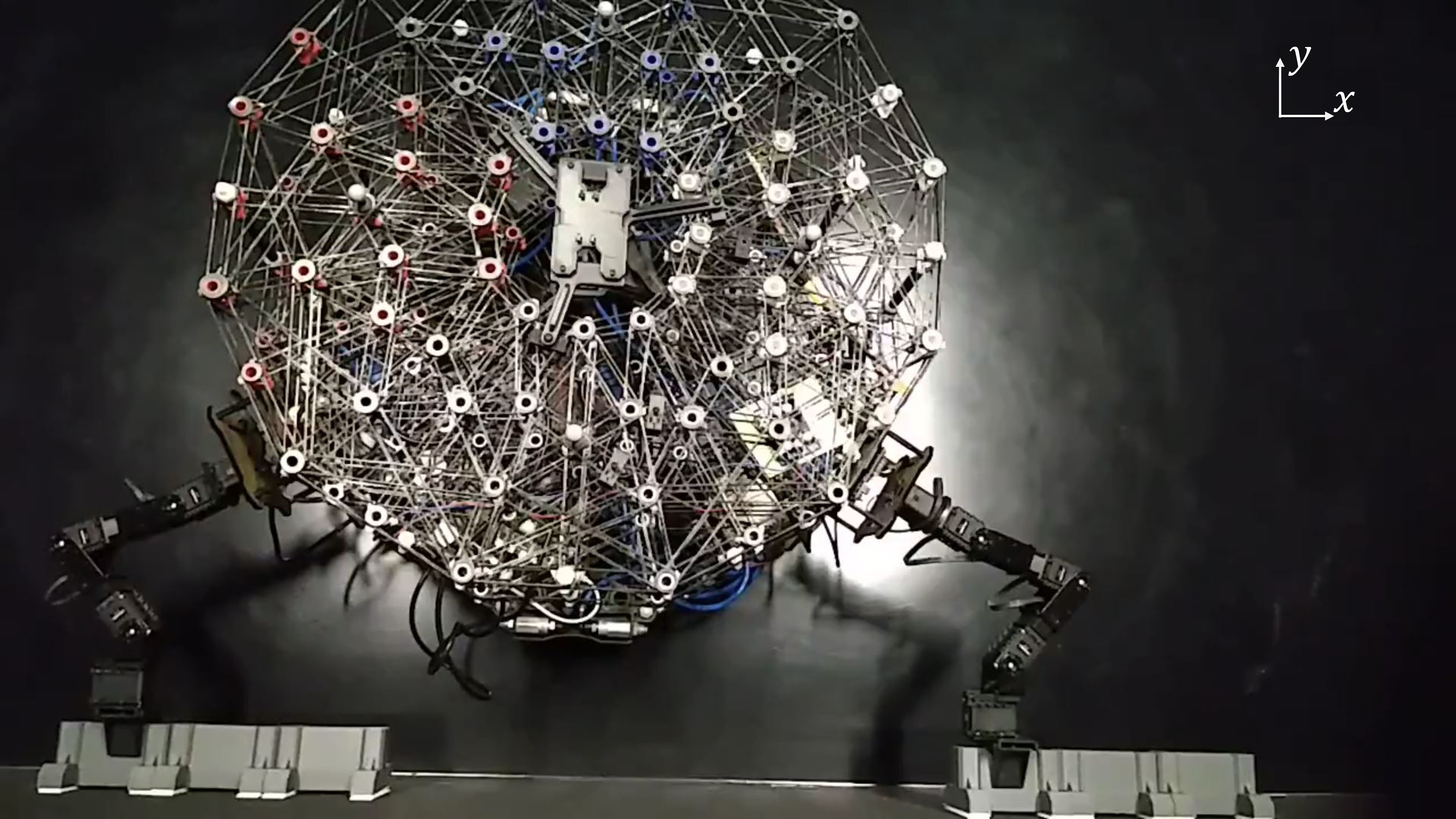}
\label{fig:exp_RAMP0}
}
\hspace{-1mm}
\subfigure[$t=20$~s]
{\includegraphics[width=0.48\linewidth]{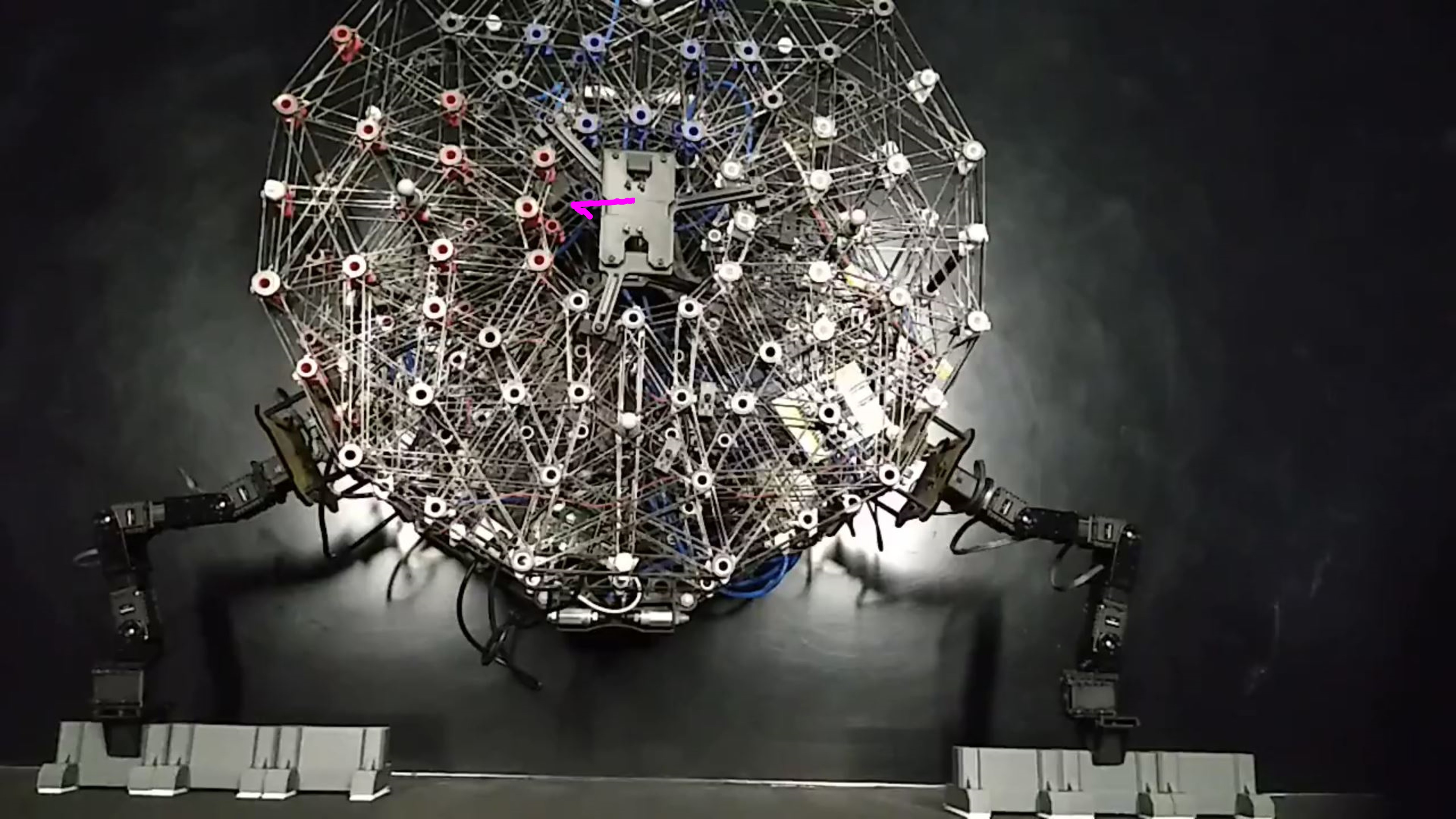}
\label{fig:exp_RAMP20}
}\\
\vspace{-2.5mm}
\subfigure[$t=38$~s]
{\includegraphics[width=0.48\linewidth]{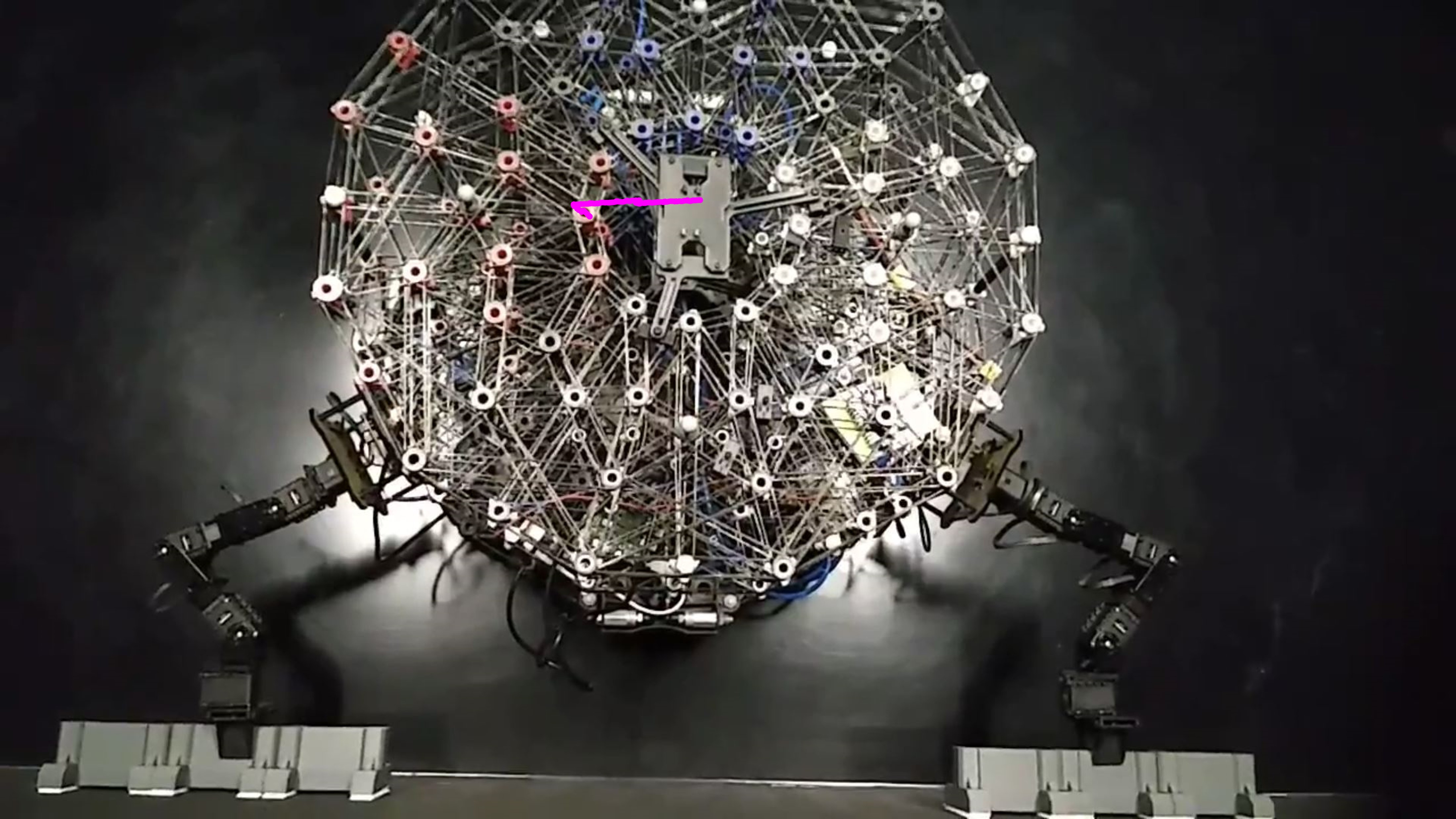}
\label{fig:exp_RAMP38}
}
\caption{Snapshots of experiments}
\label{fig:exp_RAMP}
\end{figure}

Fig.~\ref{fig:exp_BL} and Fig.~\ref{fig:exp_RAMP} show the snapshots of the two experiments performed, using the baseline mobility and the new proposed mobility strategy for multi-limbed robots for asteroid exploration, respectively. The result for the baseline case shows that the robot fails to walk one cycle, detaching from the ground surface, similar to the simulation study presented in the previous section. As the motion reactions generated by the non-optimized swinging movement are excessive, the robot fails to grasp its target, failing to achieve the desired locomotion. The motion with the proposed reaction-aware planning reduces the reactions, completing the walking cycle while avoiding the slippage of the grippers.

\begin{figure}[!t]
\centering
\subfigure[Linear Acceleration]
{\includegraphics[width=0.48\linewidth]{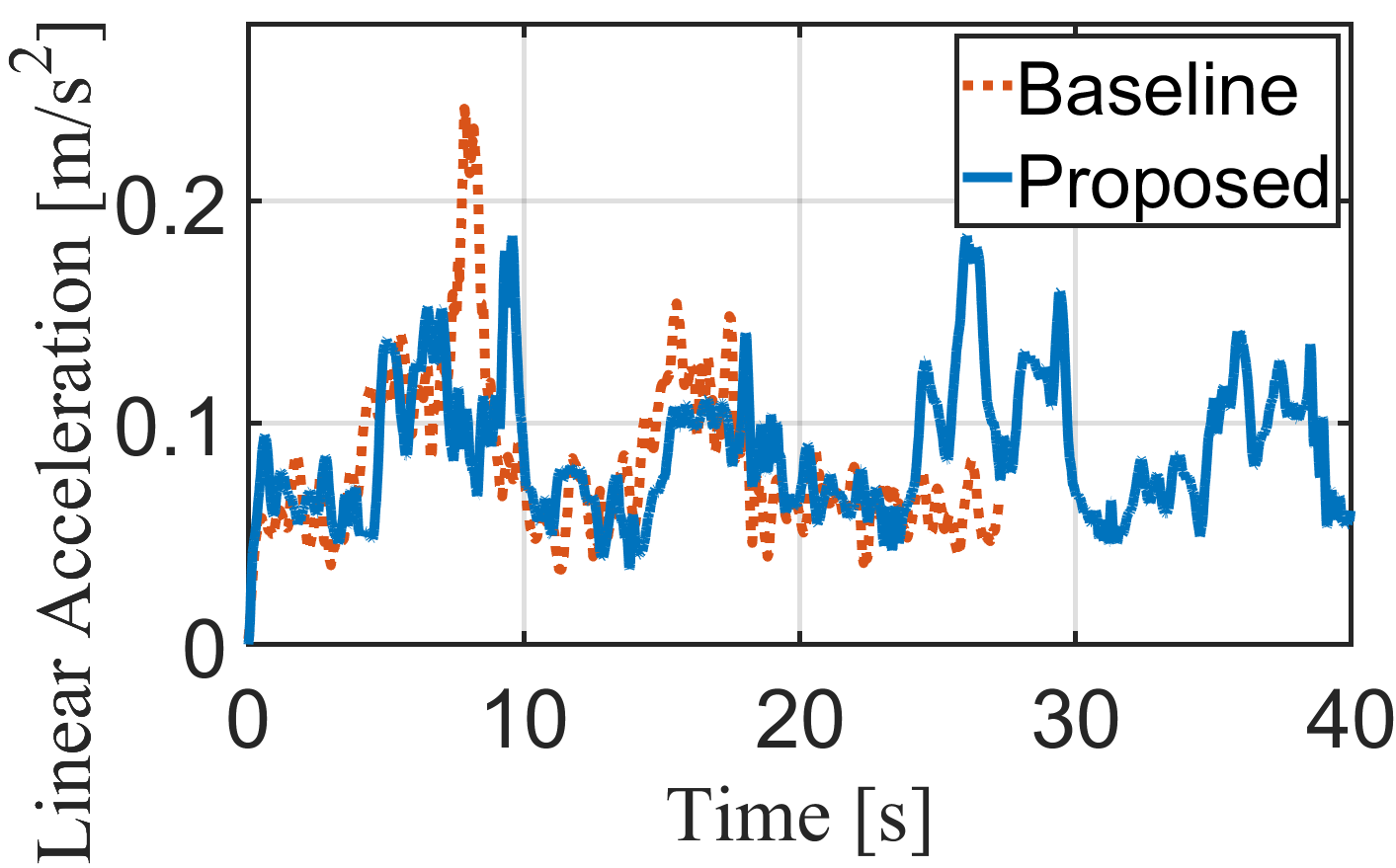}
\label{fig:lin}
}
\hspace{-0.5mm}
\subfigure[Angular Acceleration]
{\includegraphics[width=0.48\linewidth]{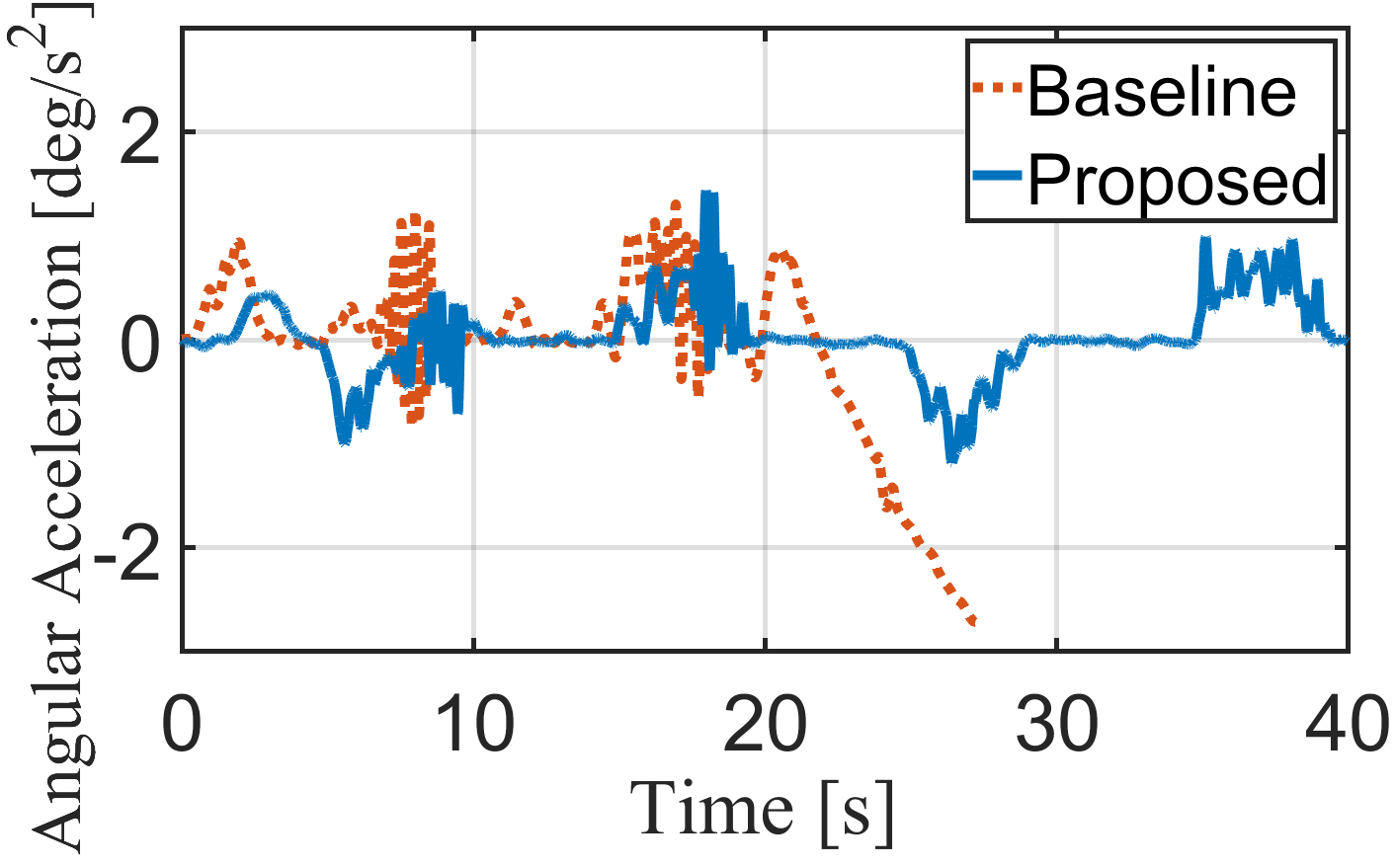}
\label{fig:ang}
}
\caption{Comparison graphs between experiments with and without the proposed strategy for mobility on asteroids.}
\label{fig:exp_graph}
\end{figure}

Fig.~\ref{fig:exp_graph} compares linear and angular acceleration data of the robot's base for both experiments obtained from an IMU sensor. We observe a smaller maximum value for both graphs as the robot uses our proposed method, indicating that it can decrease the reactions caused by the movement of a robot in microgravity conditions.

\section{Conclusions}

The method we proposed to improve the mobility of multi-limbed robots on the surface of asteroids is based on two main characteristics. The first is the gait planning that avoids collision with the ground while decreasing the chances of kinematic singularities, adjusting the base pose to the surface conditions. The second is the Reaction-Aware Motion Planning which reduces reactions, mitigating the risk of gripper slippage, with an automatic selection of the momentum distribution factor based on the manipulability index. 

This work validates the proposed mobility strategy using dynamic simulations and experiments with a system that emulates microgravity. The suggested method performed better in both cases, avoiding robot flotation and ground collisions for safer locomotion.

%


\begin{thebibliography}{21}

\bibitem {Badescu2013}
Badescu, V.: Asteroids: {Prospective} energy and material resources. Springer Science \& Business Media (2013)

\bibitem {yoshida2008}
Yoshida, K., Wilcox, B.: Space Robots and Systems. In: Siciliano, B., Khatib, O. (eds) Springer Handbook of Robotics (2008)

\bibitem {yano2006touchdown}
Yano, H. et al.: Touchdown of the {Hayabusa} spacecraft at the {Muses Sea on Itokawa}. Science. 312(5778), 1350--1353 (2006)

\bibitem {watanabe2017hayabusa2}
Watanabe, S., Tsuda, Y., Yoshikawa, M., Tanaka, S., Saiki, T., Nakazawa, S.: Hayabusa2 mission overview. Space Science Reviews. 208(1), 3--16 (2017)

\bibitem {reichert2023dart}
Reichert, S.: DART hits the bullseye. Nature Physics. 19(4), 471 (2023)

\bibitem {michikami2021boulder}
Michikami, T., Hagermann, A.: Boulder sizes and shapes on asteroids: a comparative study of {Eros, Itokawa and Ryugu}. Icarus. 357, 114282 (2021)

\bibitem {yoshimitsu2022asteroid}
Yoshimitsu, T., Kubota, T.: Asteroid surface exploration by Minerva-II small rovers. In: 18th Annual Meeting of the Asia Oceania Geosciences Society (AOGS 2021), pp. 180--182. (2022)

\bibitem {yoshida2002novel}
Yoshida, K., Maruki, T., and Yano, H.: A novel strategy for asteroid exploration with a surface robot. In: 34th COSPAR Scientific Assembly, p. 1966. (2002)

\bibitem {parness2013gravity}
Parness, A. et al.: Gravity-independent rock-climbing robot and a sample acquisition tool with microspine grippers. J. of Field Robotics. 30(6), 897--915 (2013)

\bibitem {parness2017lemur}
Parness, A. et al.: Lemur 3: A limbed climbing robot for extreme terrain mobility in space. In: IEEE International Conference on Robotics and Automation, pp. 5467--5473. (2017)

\bibitem {chacin2009motion}
Chacin, M., Mora, A., Yoshida, K.: Motion control of multi-limbed robots for asteroid exploration missions. In: IEEE International Conference on Robotics and Automation, pp. 3037--3042. (2009)

\bibitem {yuguchi2016analysis}
Yuguchi, Y., Ribeiro, W.F.R., Nagaoka, K., Yoshida, K.: Analysis on motion control based on reaction null space for ground grip robot on an asteroid. Transactions of the Japan Society for Aeronautical and Space Sciences, Aerospace Technology Japan. 14(ists30), Pk\_125--Pk\_130 (2016)

\bibitem {uno2019gait}
Uno, K. et al.: Gait planning for a free-climbing robot based on tumble stability. In: IEEE/SICE International Symposium on System Integration, pp. 289--294. (2019)

\bibitem {kato2022pin}
Kato, T., Uno, K., Yoshida, K.: A pin-array structure for gripping and shape recognition of convex and concave terrain profiles. In: IEEE International Conference on Robotics and Biomimetics, pp. 1365--1370. (2022)

\bibitem {chen2022adaptive}
Chen, J., Xu, K., Ding, X.: Adaptive gait planning for quadruped robot based on center of inertia over rough terrain. Biomimetic Intelligence and Robotics. 2(1), 100031 (2022)

\bibitem {ribeiro2023ramp}
Ribeiro, W.F.R., Uno, K., Imai, M., Murase, K., Yoshida, K.: RAMP: Reaction-Aware Motion Planning of Multi-Legged Robots for Locomotion in Microgravity. In: IEEE International Conference on Robotics and Automation, pp. 11845--11851. (2023)

\bibitem {ribeiro2022low}
Ribeiro, W.F.R., Uno, K., Yoshida, K.: Low-Reaction Trajectory Generation for a Legged Robot in Microgravity. In: IEEE/SICE International Symposium on System Integration, pp. 505--510. (2022)

\bibitem {uno2022climblab}
Uno, K. et al.: Climblab: MATLAB simulation platform for legged climbing robotics. Robotics for Sustainable Future: CLAWAR 2021, pp. 229--241. (2022)

\bibitem {uno2021hubrobo}
Uno, K. et al.: Hubrobo: a lightweight multi-limbed climbing robot for exploration in challenging terrain. In: IEEE 20th International Conference on Humanoid Robots, pp. 209--215. (2021)

\bibitem {ribeiro2020dynamic}
Ribeiro, W.F.R., Uno, K., Nagaoka, K., Yoshida, K.: Dynamic equilibrium of climbing robots based on stability polyhedron for gravito-inertial acceleration. In: 23rd International Conference on Climbing and Walking Robots and the Support Technologies for Mobile Machines, pp. 297--304. (2020)

\bibitem {yalcin2023ultra}
Yal\c{c}{\i}n, B.C., Martinez, C., Coloma, S., Skrzypczyk, E., Olivares-Mendez, M.: Ultra Light Floating Platform: An Orbital Emulator for Space Applications. In: IEEE ICRA 2023 Late Breaking Results Poster Presentation. (2023)

\end{thebibliography}
\end{document}